\documentclass[runningheads]{llncs}
\usepackage[T1]{fontenc}
\usepackage{graphicx,verbatim}
\usepackage{array}       
\usepackage{arydshln}    
\usepackage{multirow}    
\usepackage{graphicx}    

\usepackage[table]{xcolor}
\usepackage{hyperref}

\hypersetup{
    colorlinks,
    linkcolor={red!50!black},
    citecolor={blue!50!black},
    urlcolor={red!50!black}
}

\begin{document}
%
\title{Subgroup performance analysis of adaptation strategies for chest X-ray foundation models}
\titlerunning{Subgroup performance analysis of adaptation strategies for CXR FMs}
%
\author{Dhruv Gupta\inst{1} 
\and
Emma A. M. Stanley\inst{1} 
\and
Fabio De Sousa Ribeiro\inst{1,2} 
\and
Sujal R. Desai\inst{3,4} 
\and
Ben Glocker\inst{1,2}
}
%

\authorrunning{D. Gupta et al.}
%
\institute{Department of Computing, Imperial College London, UK
\and
Causality in Healthcare AI Hub, UK
\and
Department of Radiology, Royal Brompton Hospital, UK
\and
National Heart \& Lung Institute, Imperial College London, UK\\
\email{d.gupta25@imperial.ac.uk}}

%
  
\maketitle              
\begin{abstract}
Foundation models are increasingly adapted for downstream medical imaging tasks, yet the influence of the chosen adaptation strategy on subgroup fairness remains poorly understood. We investigate how three parameter-efficient adaptation techniques, including linear heads on the raw CLS token, an MLP, and an attention-pooling module over multi-layer patch features, affect both pathology classification performance and subgroup disparities when applied to the frozen Rad-DINO chest X-ray encoder. Using MIMIC-CXR, we evaluate eight pathologies across race, sex, and imaging-view subgroups on a prevalence-preserving, demographically balanced test set, and additionally probe how strongly each adapter encodes protected attributes. We find that attention pooling achieves the strongest overall discriminative performance and encodes attributes, particularly race, most strongly, but that improved overall performance does not consistently reduce subgroup disparities. Notably, stronger attribute encoding did not correspond to larger disparities: early network layers encoded race most weakly yet produced the largest subgroup performance gaps. Exploring different attention-pooling layer combinations further revealed no consistent relationship between the layers pooled, attribute encoding strength, and subgroup fairness. Our results indicate that richer, more expressive representations can improve accuracy while leaving fairness implications task-dependent and unpredictable, which must be assessed directly and per-task rather than inferred from encoding strength or overall performance alone.

\keywords{Subgroup fairness \and foundation models \and bias analysis.}

\end{abstract}

\section{Introduction}

Medical imaging AI models have been shown to reproduce biases present in their training data, particularly for historically underserved demographics such as female and Black patients \cite{seyyed-kalantari_underdiagnosis_2021}. Such biases are of particular concern given the ability of deep learning models to learn and encode strong representations for protected characteristics such as race and biological sex \cite{gichoya_ai_2022}, raising the risk that this information could be exploited as a shortcut for the primary diagnostic task, potentially amplifying subgroup disparities \cite{glocker_algorithmic_2023}.

As the adoption of large scale foundation models (FMs) has grown due to their generalisation capabilities and often state-of-the-art performance when fine-tuned or adapted for downstream tasks, fairness analysis of these models has become particularly important. Studies have shown that subgroup performance disparities exist even when using various FMs across medical image modalities \cite{jin_fairmedfm_2024}, and that FMs can encode protected characteristics more strongly than task-specific models \cite{glocker_risk_2023}. This suggests that classifiers built on top of FM feature embeddings may be susceptible to inheriting or exploiting these encoded characteristics, resulting in biased outcomes across relevant subgroups.

\begin{figure}[t]
\includegraphics[width=\textwidth]{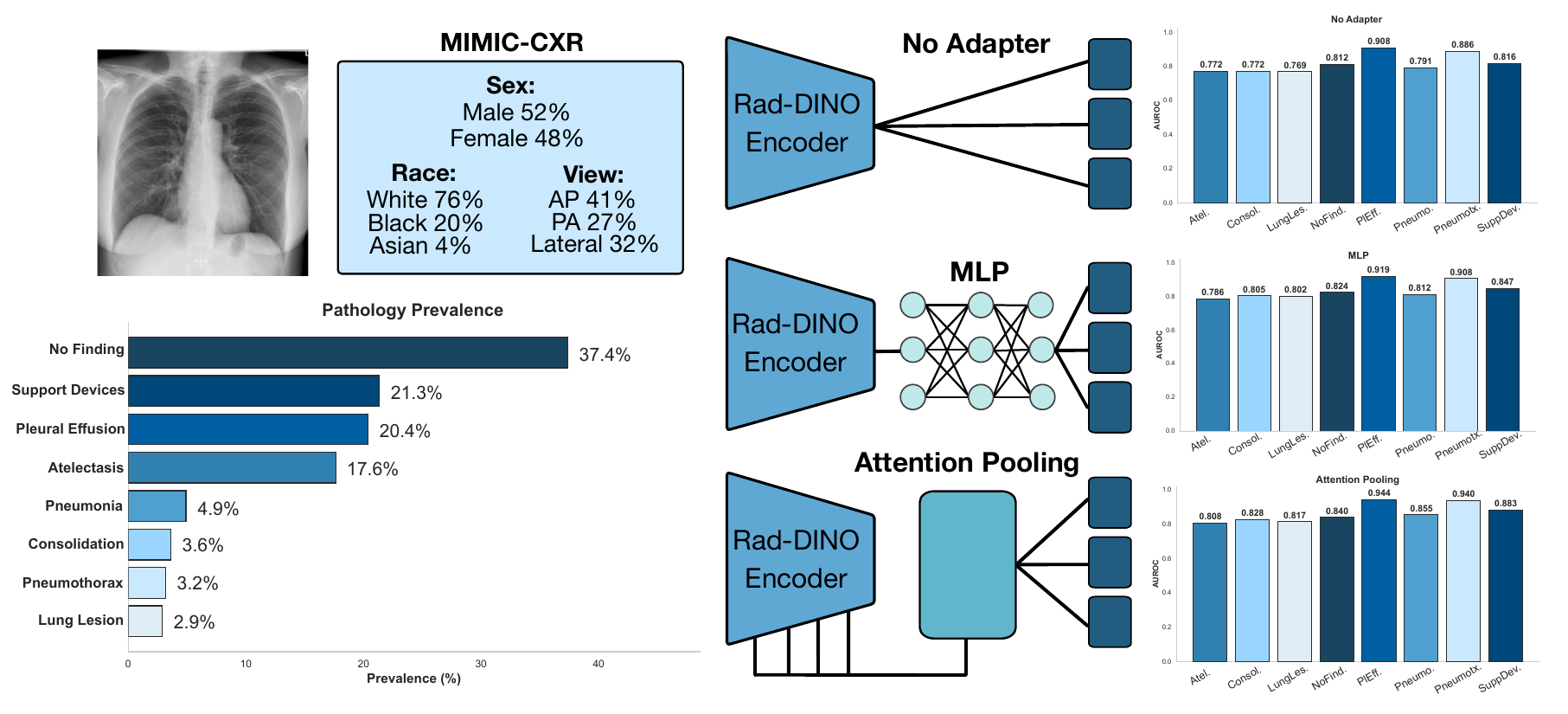}
\caption{Overview of the dataset and adapter architectures used to evaluate demographic and imaging-view bias on top of the frozen Rad-DINO foundation model backbone. Three adapter variants are considered, all paired with linear decision heads for downstream pathology classification: 1) \texttt{No Adapter}, using the 768-dimensional CLS token from the final encoder layer directly; 2) an \texttt{MLP} applied to the same CLS token; and 3) an \texttt{Attention Pooling} adapter over features concatenated from four transformer layers across the encoder. Overall AUROC per pathology is reported on the resampled test set used throughout this work.} \label{overview_fig}
\end{figure}

Different strategies for adapting FM features for downstream tasks exist, but the effect of a particular adaptation strategy on subgroup performance is not well understood. Due to computational constraints and the risk of catastrophic forgetting (the trade-off between updating more parameters to improve task-specific fit versus preserving the generalisability of pre-trained representations) associated with full model fine-tuning, adaptation techniques commonly leverage the representations of a pre-trained frozen encoder for a specific task  \cite{dutt_fairtune_2024}. Such parameter-efficient fine-tuning techniques have been shown to achieve state-of-the-art performance when applied to powerful foundation models \cite{ilse_data_2025}. Recent work by Ribeiro et al. \cite{ribeiro_scaling_2026}, for example, demonstrated that discriminative performance of the Rad-DINO chest X-ray FM \cite{perez-garcia_exploring_2025} was substantially improved by using an \textit{attention pooling} model over concatenated multi-layer patch features. Interestingly, this approach appeared to improve race prediction in particular, relative to linear or MLP heads applied to the vision transformer's learned CLS token. In this case, the adapter was explicitly trained to optimise these representations for demographic subgroup prediction, however, it remains unclear how this strategy affects subgroup performance in pathology classification tasks.

In this work, we explore how different parameter-efficient adapter methods affect the downstream performance of Rad-DINO \cite{perez-garcia_exploring_2025}, together with subgroup performance across race, sex, and imaging view (frontal vs. lateral). In addition, we investigate how the choice of attention-pooled layers affects subgroup encoding and performance, showing that neither strong overall performance nor strong attribute encoding reliably predicts fair subgroup outcomes.

\section{Methods}

\subsection{Models and Training}

To assess different strategies for adapting the frozen Rad-DINO encoder to multi-label pathology classification, we compare three adapters, each followed by a linear decision head per label class, 1) \texttt{No Adapter}, where the linear heads act directly on the final layer CLS token, 2) \texttt{MLP}, a hidden-layer projection of the CLS token with non-linearity, and 3) \texttt{Attention Pooling}. All adapters were trained jointly with the decision heads on top of the frozen encoder backbone to minimise binary cross-entropy loss across eight pathology classification tasks (see Fig. \ref{overview_fig}). We compare these three approaches to assess how increasing adapter complexity affects downstream pathology classification performance and subgroup performance disparities. We also evaluate the strength of subgroup attribute encoding (i.e., predictive performance for sex, race, and view classification) to investigate whether more expressive adapters either exploit or suppress spurious associations present in the FM embeddings.

Following \cite{ribeiro_scaling_2026}, the attention pooling adapter mechanism evaluated in this work takes advantage of the Rad-DINO ViT-B backbone, by extracting the CLS and patch tokens from four different attention blocks across the model and concatenating them along the feature dimension. This results in a 4 $\times$ 768 dimensional representation. A learned query vector, one per pathology task, then attends over this concatenated token sequence via multi-head attention, producing a single pooled representation per task. The per-task linear decision heads then classify across the logits from the pooled output. The query vectors are task-specific, but the attention weights themselves are shared across all tasks, and are only optimised by the pathology classification losses in our setup.

Prior work applying attention pooling to Rad-DINO \cite{ilse_data_2025,ribeiro_scaling_2026} selected which layers to pool from heuristically. Since different layers in deep learning models have been shown to encode different information \cite{bolya_perception_2025}, we explore different layer combinations, evaluating each on overall and subgroup pathology performance, as well as on the strength of attribute encoding. We considered the following four layer combinations Early Layers (2,3,4,5), Late Layers (9,10,11,12), Split Layers (2,3,11,12) and Even Layers (3,6,9,12). Even Layers (3,6,9,12) matches the configuration used in prior work \cite{ribeiro_scaling_2026}, and is the configuration used for the primary comparison against \texttt{No Adapter} and \texttt{MLP}.

Following convergence on the eight pathology tasks, separate classification probes for race, sex, and view were trained on top of the adapters to determine the extent to which the attribute subgroups were encoded in the learned representations. For \texttt{No Adapter} and \texttt{MLP}, this takes the standard form of a linear probe where a new linear classifier is fit directly on the frozen adapter's output. Since the \texttt{Attention Pooling} configurations have no fixed embedding to probe, we therefore introduce new, randomly initialised query vectors and linear heads for race, sex, and view, and train only these new parameters whilst the shared attention weights optimised during pathology training remain frozen, receiving no gradient from the attribute objective. 

\subsection{Dataset}

We investigate fairness across demographic and imaging view subgroups using the MIMIC Chest X-Ray dataset (MIMIC-CXR), comprising a train/test split of 332,249/33,394 images \cite{johnson_mimic-cxr_2019}. Clinical pathology labels were extracted from the findings section of radiology reports using CheXbert, yielding CheXpert-14-compatible pathology labels, following the procedure described in \cite{ribeiro_scaling_2026}. Patient `Sex', `Age', and acquisition `View' information were extracted from the available metadata, and patient `Race' labels were consolidated into `Asian', `Black', and `White' categories. From the 14 available CheXpert pathology labels, a subset of eight was used in our evaluation. When training and evaluating different adapters, all non-positive cases were considered to be negative, including cases where labels are potentially uncertain, aligning with prior work \cite{glocker_algorithmic_2023}.

\subsection{Evaluation}
We evaluate all task adaptation methods using the Area Under the Receiver Operating Characteristic Curve (AUROC), which provides a threshold-independent measure of discriminative performance of the model. To enable valid and comparable performance accounting for relative size and disease prevalence across subgroups, we adopt test set rebalancing for pathology classification~\cite{glocker_algorithmic_2023}. This procedure constructs balanced evaluation subsets through resampling to achieve equal representation across demographic subgroups while maintaining disease prevalence. This ensures that differences in subgroup AUROC can be attributed to model behaviour rather than a prevalence skew in the dataset. When comparing subgroup performance, we compute the difference in AUROC between each subgroup and the overall AUROC computed on the full evaluation cohort for that task, stratified by individual attribute subgroups. To determine attribute encoding, we compute one-vs-rest (OvR) AUROC for each class, using the test set. Performance values are reported with 95\% confidence intervals based on bootstrap resampling with replacement (100 iterations).

\section{Results}

\subsection{Impact of Adapter Choice on Downstream Task Performance}

Table \ref{tab_global_metrics} reports the performance across the three different adapters on the eight different pathology tasks. Across all tasks, the best performance results are obtained using \texttt{Attention Pooling}, although the \texttt{MLP} also improves performance consistently over the \texttt{No Adapter} baseline where linear decision heads were trained on the raw CLS token outputs. The performance improvements of \texttt{Attention Pooling} can potentially be attributed to the larger adapter capacity, requiring 37,810,184 parameters compared to 598,280 for the \texttt{MLP}, as well as the expressiveness of the features extracted from earlier Rad-DINO layers.

In addition to the primary task of pathology classification, the strength of attribute encoding is reported in Table \ref{tab_global_metrics}, by means of AUROC based on the frozen final adapter layers and query vectors. \texttt{Attention Pooling} shows the strongest degree of attribute encoding, with a notable increase in race prediction, particularly in the Asian subgroup. Encoding of Race (and Sex, to a lesser degree) is higher with \texttt{No Adapter} compared to the \texttt{MLP}, which could reflect representations optimised for pathology prediction that lose some degree of attribute encoding from the base-level foundation model.


\newcommand{\ci}[2]{{\tiny [#1, #2]}}

\begin{table}[htbp]
\caption{Overall AUROC performance on eight pathologies across three adapters trained on top of the frozen Rad-DINO backbone: \texttt{No Adapter} (linear heads on the raw encoder output), \texttt{MLP}, and \texttt{Attention Pooling} (layers 3, 6, 9, 12). Bold indicates the highest point estimate per row.}\label{tab_global_metrics}
\centering
\begin{tabular*}{\textwidth}{@{\extracolsep{\fill}}llccc@{}}
\hline
\textbf{Cat.} & \textbf{Class} & \textbf{No Adapter} & \textbf{MLP} & \textbf{Attention Pooling} \\
\hline
\multirow{8}{*}{\textbf{Path.}}
& Atelectasis      & 0.772 \ci{0.768}{0.782} & 0.786 \ci{0.782}{0.795} & \textbf{0.808} \ci{0.803}{0.815} \\
& Consolidation    & 0.772 \ci{0.764}{0.790} & 0.805 \ci{0.799}{0.823} & \textbf{0.828} \ci{0.823}{0.843} \\
& Lung Lesion      & 0.769 \ci{0.764}{0.791} & 0.802 \ci{0.795}{0.816} & \textbf{0.817} \ci{0.811}{0.835} \\
& No Finding       & 0.812 \ci{0.806}{0.814} & 0.824 \ci{0.818}{0.826} & \textbf{0.840} \ci{0.835}{0.842} \\
& Pleural Effusion & 0.908 \ci{0.902}{0.910} & 0.919 \ci{0.915}{0.922} & \textbf{0.944} \ci{0.941}{0.947} \\
& Pneumonia        & 0.791 \ci{0.781}{0.799} & 0.812 \ci{0.807}{0.823} & \textbf{0.855} \ci{0.850}{0.864} \\
& Pneumothorax     & 0.886 \ci{0.880}{0.900} & 0.908 \ci{0.899}{0.918} & \textbf{0.940} \ci{0.926}{0.945} \\
& Support Devices  & 0.816 \ci{0.809}{0.822} & 0.847 \ci{0.840}{0.853} & \textbf{0.883} \ci{0.877}{0.888} \\
\hdashline
\multirow{7}{*}{\textbf{Attr.}}
& Race: Asian      & 0.876 \ci{0.862}{0.887} & 0.771 \ci{0.755}{0.785} & \textbf{0.943} \ci{0.936}{0.949} \\
& Race: Black      & 0.893 \ci{0.887}{0.897} & 0.809 \ci{0.802}{0.816} & \textbf{0.946} \ci{0.942}{0.949} \\
& Race: White      & 0.890 \ci{0.885}{0.894} & 0.831 \ci{0.825}{0.835} & \textbf{0.945} \ci{0.942}{0.948} \\
& Sex: Male        & 0.997 \ci{0.997}{0.998} & 0.989 \ci{0.988}{0.989} & \textbf{0.999} \ci{0.999}{0.999} \\
& View: AP         & 0.999 \ci{0.999}{1.000} & 0.999 \ci{0.999}{1.000} & \textbf{1.000} \ci{1.000}{1.000} \\
& View: PA         & 0.998 \ci{0.997}{0.998} & 0.997 \ci{0.996}{0.998} & \textbf{0.999} \ci{0.998}{0.999} \\
& View: Lateral    & 0.999 \ci{0.998}{0.999} & 0.999 \ci{0.998}{0.999} & \textbf{1.000} \ci{0.999}{1.000} \\
\hline
\end{tabular*}
\end{table}

\subsection{Subgroup Performance Disparities from Different Adapters}

\begin{figure}[t]
\includegraphics[width=\textwidth]{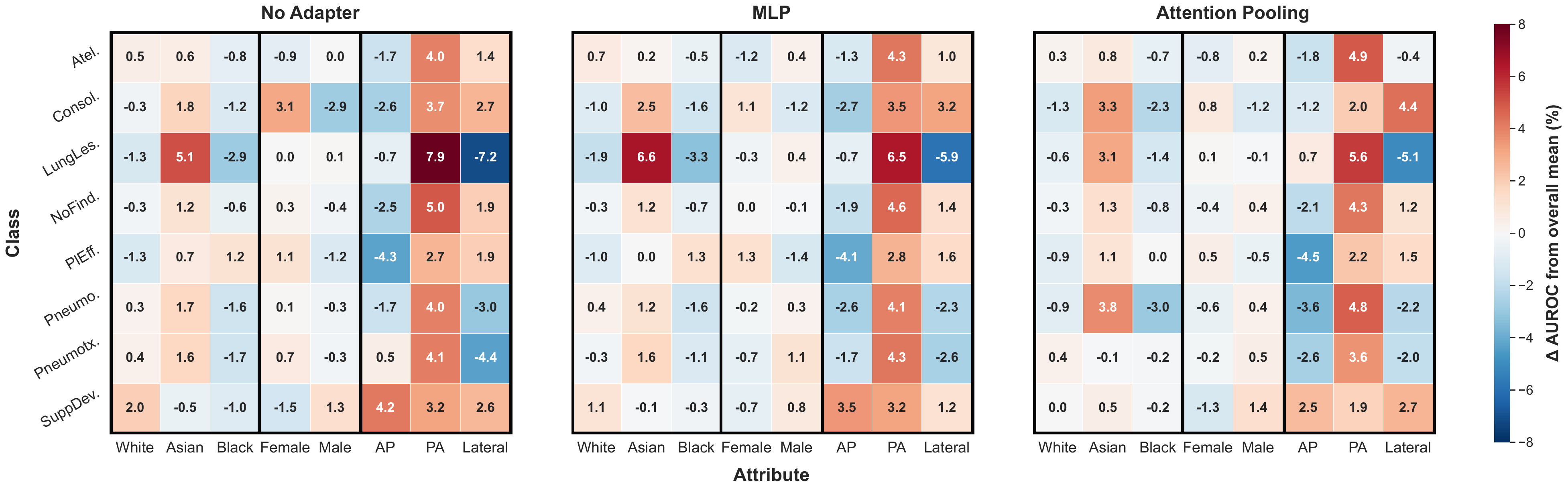}
\caption{Subgroup performance characteristics presented across the three different adapter types. Values are reported as the percentage difference in AUROC from the mean for each pathology task. Columns are subdivided into attribute groups.} \label{subgroup_heatmap}
\end{figure}

Subgroup disparities relative to the overall model AUROC for each pathology are visualised in Figure \ref{subgroup_heatmap}.
Generally, the choice of adapter does not consistently impact the degree of subgroup disparities observed within the various pathologies. For instance, \texttt{Attention Pooling} increases the performance disparities over \texttt{No Adapter} in some categories (e.g., View-Atelectasis), but reduces disparities for others (e.g., Race-Lung Lesion). While the model with \texttt{No Adapter} exhibits the most extreme disparities (particularly within the Race and View attributes), \texttt{Attention Pooling} can minimise these extremes but can introduce consistently higher performance disparities across subgroups (e.g., Race-Pneumonia). Only the View attribute exhibits a relatively consistent trend of decreasing disparities when changing from \texttt{No Adapter} to \texttt{MLP} to \texttt{Attention Pooling}. We also note that the subgroup performance for some tasks remains relatively consistent regardless of the adapter method (e.g., Pleural Effusion, Pneumothorax), while each having very different degrees of prevalence in the dataset (see Fig. \ref{overview_fig}).

\subsection{Effect of Different Attention Pooling Layer Combinations}

\begin{table}[htbp]
\caption{Overall AUROC performance across four attention-pooling layer configurations, evaluated on the resampled test set (pathology) and the standard test set (attribute). Bold indicates the highest point estimate per row.}\label{tab_attn_layer_metrics}
\centering
\begin{tabular*}{\textwidth}{@{\extracolsep{\fill}}llcccc@{}}
\hline
\textbf{Cat.} & \textbf{Class} & \textbf{Early} & \textbf{Late} & \textbf{Split} & \textbf{Even} \\
\hline
\multirow{8}{*}{\textbf{Path.}}
& Atelectasis      & 0.797\ci{0.792}{0.804} & 0.806\ci{0.801}{0.813} & 0.804\ci{0.800}{0.811} & \textbf{0.808}\ci{0.803}{0.815} \\
& Consolidation    & 0.829\ci{0.826}{0.846} & \textbf{0.831}\ci{0.824}{0.845} & 0.826\ci{0.821}{0.843} & 0.828\ci{0.823}{0.843} \\
& Lung Lesion      & 0.814\ci{0.810}{0.832} & \textbf{0.824}\ci{0.816}{0.839} & 0.813\ci{0.804}{0.831} & 0.817\ci{0.811}{0.835} \\
& No Finding       & 0.831\ci{0.826}{0.834} & \textbf{0.841}\ci{0.836}{0.843} & 0.837\ci{0.831}{0.839} & 0.840\ci{0.835}{0.842} \\
& Pleural Effusion & 0.942\ci{0.938}{0.944} & 0.944\ci{0.941}{0.946} & 0.943\ci{0.940}{0.945} & \textbf{0.944}\ci{0.941}{0.947} \\
& Pneumonia        & 0.839\ci{0.830}{0.847} & 0.852\ci{0.845}{0.859} & 0.842\ci{0.838}{0.851} & \textbf{0.855}\ci{0.850}{0.864} \\
& Pneumothorax     & 0.936\ci{0.923}{0.940} & 0.942\ci{0.930}{0.950} & \textbf{0.944}\ci{0.931}{0.948} & 0.940\ci{0.926}{0.945} \\
& Support Devices  & 0.869\ci{0.862}{0.873} & 0.883\ci{0.876}{0.887} & 0.880\ci{0.874}{0.885} & \textbf{0.883}\ci{0.877}{0.888} \\
\hdashline
\multirow{7}{*}{\textbf{Attr.}}
& Race: Asian      & 0.864\ci{0.853}{0.872} & \textbf{0.954}\ci{0.948}{0.960} & 0.939\ci{0.933}{0.946} & 0.943\ci{0.936}{0.949} \\
& Race: Black      & 0.880\ci{0.875}{0.886} & \textbf{0.951}\ci{0.948}{0.954} & 0.945\ci{0.942}{0.949} & 0.946\ci{0.942}{0.949} \\
& Race: White      & 0.877\ci{0.872}{0.882} & \textbf{0.950}\ci{0.947}{0.953} & 0.944\ci{0.940}{0.946} & 0.945\ci{0.942}{0.948} \\
& Sex: Male        & 0.987\ci{0.986}{0.988} & \textbf{0.999}\ci{0.999}{0.999} & 0.999\ci{0.999}{0.999} & 0.999\ci{0.999}{0.999} \\
& View: AP         & 1.000\ci{0.999}{1.000} & \textbf{1.000}\ci{1.000}{1.000} & 1.000\ci{1.000}{1.000} & 1.000\ci{1.000}{1.000} \\
& View: PA         & 0.999\ci{0.998}{0.999} & \textbf{0.999}\ci{0.998}{0.999} & 0.999\ci{0.998}{0.999} & 0.999\ci{0.998}{0.999} \\
& View: Lateral    & 0.999\ci{0.999}{0.999} & \textbf{1.000}\ci{0.999}{1.000} & 1.000\ci{0.999}{1.000} & 1.000\ci{0.999}{1.000} \\
\hline
\end{tabular*}
\end{table}

\begin{figure}
\includegraphics[width=\textwidth]{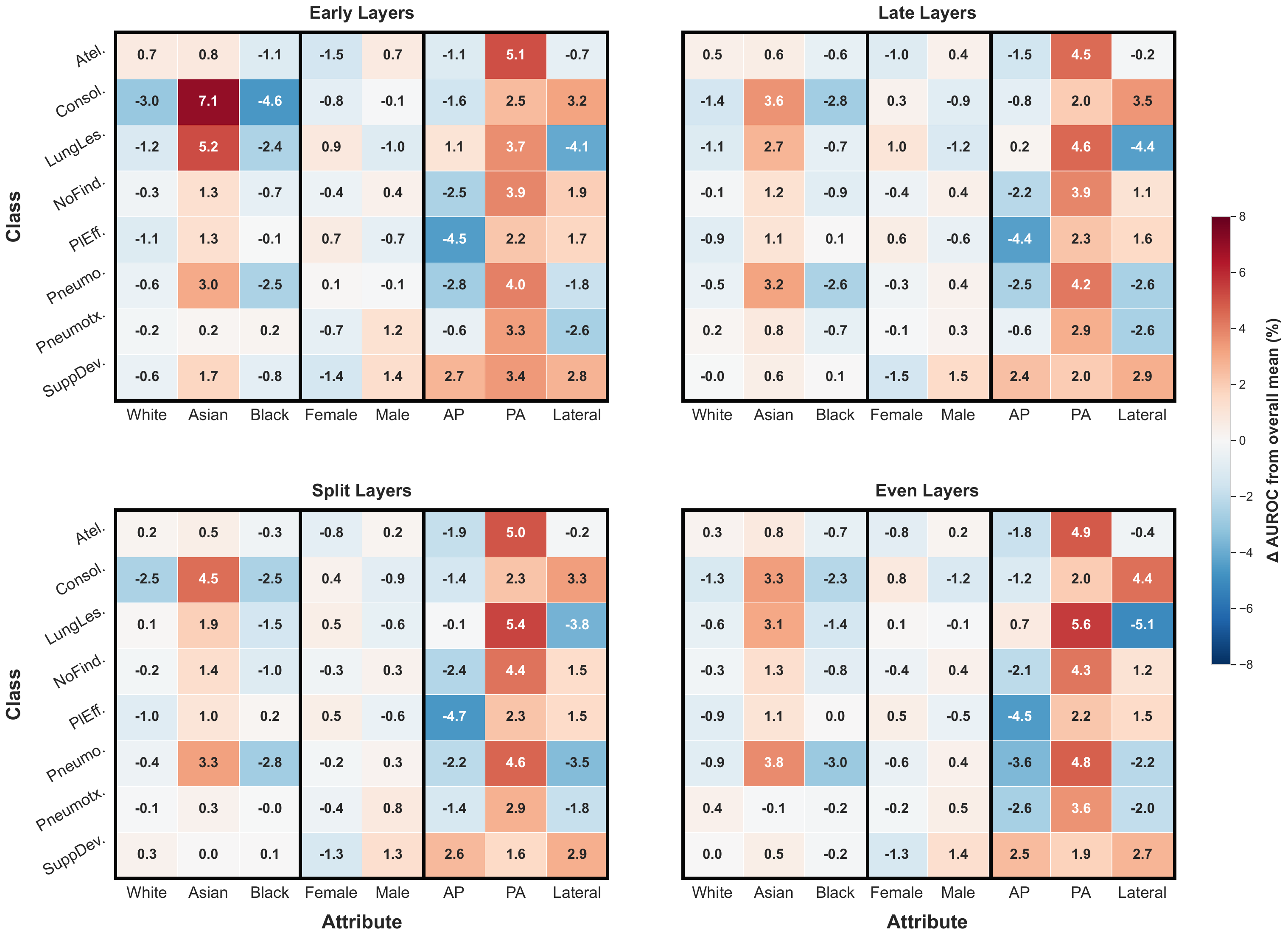}
\caption{Subgroup performance characteristics across four different layer combinations used for the attention pooling adapter on the Rad-DINO backbone. Values are reported as percentage AUROC differences from the mean across each pathology. } \label{attention_subgroup}
\end{figure}


We also explore the impact on overall and subgroup performance when using different \texttt{Attention Pooling} layer combinations. Table \ref{tab_attn_layer_metrics} shows that these different combinations can result in varying performance outcomes across tasks. Lung Lesion and Pneumonia classification performance is sensitive to the choice of pooled layers, while Atelectasis, No Finding, and Pleural Effusion remain relatively consistent. These stable pathologies also exhibit relatively low variation in subgroup performance across configurations (see Fig. \ref{attention_subgroup}). Conversely, the overall performance for Consolidation is relatively invariant to layer selection, however, subgroup performance shows noticeable differences.

For attribute encoding, Race is notably represented strongest in the later layers of Rad-DINO (i.e., layers 9,10,11,12). All three configurations (Late, Split and Even) which incorporate later layers encode race more strongly than the Early layer configuration (i.e., layers 2,3,4,5). In contrast, Sex and View attributes are strongly encoded even in the early layers with 0.987 and $\approx$1.000 AUROC, respectively. This stronger attribute encoding in the later layers does not correspond to more consistent subgroup performance, with the largest variations in subgroup performance occurring in the Early and Even layer configurations (e.g., Race-Consolidation and View-Lung Lesion).

\section{Discussion \& Conclusions}
We conducted a comprehensive bias analysis and found that foundation model adapters which improve overall downstream performance can impact subgroup disparities, but do not consistently reduce them. This fairness–utility trade-off, also reported elsewhere \cite{jin_fairmedfm_2024}, reinforces that such trade-offs must be assessed per application, since no single method is likely to deliver both robust and fair performance across all tasks. The \texttt{No Adapter} and \texttt{MLP} cases exhibited similar disparity patterns, differing mainly in magnitude across tasks. This is likely due  to both adapters only relying on Rad-DINO's final layer CLS embeddings. \texttt{Attention Pooling} instead draws on CLS and patch tokens from multiple layers, exploiting representations at varying levels of abstraction. Because pathologies and demographic attributes are tied to distinct imaging markers, this added expressivity plausibly enables the model to select features according to their relevance for each task—especially for FMs, whose self-supervised objectives yield generic representations that are not explicitly tuned for any specific downstream use. 

This richer representation, however, does not translate predictably into better or worse subgroup performance. Although \texttt{Attention Pooling} encoded Race far more strongly in later layers, this did not produce systematically larger disparities; in fact, the early layers encoded Race most weakly yet yielded the largest disparities. This aligns with prior work \cite{stanley_where_2025}, confirming that strong attribute encoding is not itself evidence of shortcut behaviour and must be interpreted alongside subgroup performance. Richer representations can improve accuracy, but their fairness implications must be evaluated directly and per task.

\textit{Limitations.} Our analysis uses a single dataset, so population-specific confounders may affect our findings, and other datasets or clinical settings may yield different performance and fairness characteristics. While Rad-DINO is a popular and powerful model, a similar analysis should be carried out on other FMs. We also rely on AUROC alone; a broader toolkit would more directly probe shortcut exploitation and other forms of bias.

Since pathologies and attributes appear to rely on different representations across the Rad-DINO network, future work could learn a per-task attention-pooling scheme that selects which layers to pool, rather than sharing one fixed combination. A joint objective could likewise define an explicit operating point balancing overall performance against fairness, rather than treating the two as separate, post-hoc concerns.

\textit{Reproducibility.} Code for data processing, experiments, and analysis is available at:
\url{https://github.com/dhruvg97/adapter-subgroup-analysis}.

    

\begin{credits}
\subsubsection{\ackname} This work was supported by the Royal Academy of Engineering as part of the Kheiron/RAEng Research Chair. D.G. is supported by UK Research and Innovation [UKRI AI Centre for Doctoral Training in Digital Healthcare grant number EP/Y030974/1].
E.A.M.S. acknowledges funding from Natural Sciences and Engineering Council of Canada (NSERC) Postdoctoral Research Award. F.R. and B.G. acknowledge the support of the UKRI AI programme, and the EPSRC, for CHAI-EPSRC Causality in Healthcare AI Hub (grant no. EP/Y028856/1). 

\subsubsection{\discintname}
Ben Glocker is a part-time employee of DeepHealth. Sujal Desai is Clinical Director for DMC Imaging Ltd., Tutor at Boehringer-Ingelheim ILD Academy and ad-hoc Consultant for ArgenX.
\end{credits}

%
%
%
%

\bibliographystyle{splncs04}  
\bibliography{references}






\end{document}